%% file: SoRo.tex
\documentclass[journal]{IEEEtran}

\usepackage{amsfonts,amssymb,amsbsy,latexsym,amsmath,tabulary,graphicx,times,xcolor}
\usepackage[T1]{fontenc}
\usepackage{url,multirow,morefloats,floatflt,cancel,tfrupee,multicol}
\usepackage{colortbl}
\usepackage{xcolor}
\usepackage{pifont}
\usepackage[nointegrals]{wasysym}
\usepackage{cite}
\usepackage{bm}

\title{Topology and morphology design of spherically reconfigurable homogeneous Modular Soft Robots (MSoRos)}

\author{Caitlin Freeman, Michael Maynard and Vishesh Vikas$^{1}$
\thanks{*This work is supported by the National Science Foundation under Grant No. 1830432}
\thanks{$^{1}$Caitlin Freeman, Michael Maynard and Vishesh Vikas are with the Agile Robotics Lab (ARL), University of Alabama, Tuscaloosa AL 36487, USA
        {\tt\small clfreeman@crimson.ua.edu, mcmaynard@crimson.ua.edu, vvikas@ua.edu}}%
}

\begin{document} 

\maketitle


\begin{abstract}
Imagine a swarm of terrestrial robots that can explore an environment, and, upon completion of this task, reconfigure into a spherical ball and roll out. This dimensional change alters the dynamics of locomotion and can assist them to maneuver variable terrains. The sphere-plane  reconfiguration is equivalent to projecting a spherical shell onto a plane, an operation which is not possible without distortions. Fortunately, soft materials have potential to adapt to this disparity of the Gaussian curvatures. Modular Soft Robots (MSoRos) have promise of achieving dimensional change by exploiting their continuum and deformable nature. However, design of such soft robots remains unexplored thus far. 

Here, for the first time, we present topology and morphology design of MSoRos capable of reconfiguring between spherical and planar configurations. Our approach is based in geometry, where a platonic solid determines the number of modules required for plane-to-sphere reconfiguration and the radius of the resulting sphere, e.g., four `tetrahedron-based' or six `cube-based' MSoRos are required for spherical reconfiguration. The methodology involves: (1) inverse orthographic projection of a `module-topology curve' onto the circumscribing sphere to generate the \textit{spherical topology}, (2) azimuthal projection of the spherical topology onto a tangent plane at the center of the module resulting in the \textit{planar topology}, and (3) adjusting the limb stiffness and curling ability by manipulating the geometry of cavities to realize a \textit{physical finite-width, Motor-Tendon Actuated MSoRo} that can actuate between the sphere-plane configurations. The topology design is shown to be scale invariant, i.e., scaling of base platonic solid is reflected linearly in spherical and planar topologies. The module-topology curve is optimized for the reconfiguration and locomotion ability using intramodular distortion metric that quantifies sphere-to-plane distortion. The geometry of the cavity optimizes for the limb stiffness and curling ability without compromising the actuator's structural integrity.
\end{abstract}

\input{02_Introduction}
\input{03_TopologyDesign}

\input{04_DistortionQuantification}
\input{05_Fabrication}

\section{CONCLUSIONS}
Modules of highly deformable Modular Soft Robots (MSoRos) capable of reconfiguration have potential to push the limits of adaptability and versatility by providing them with ability to change dimensions through morphing. One can visualize multiple modules of multi-limb planar soft robot morphing into a spherical ball (3D) or a millipede-like chain of soft modules (1D). However, design of such MSoRos is non-trivial and has direct effect on self-assembly and reconfiguration.

We present topology and morphology design of homogenous MSoRos based in geometry. The scale invariant topology design approach uses a base platonic solid and a module-topology curve as the design parameters. The the number of faces of the platonic solid correspond to the number of MSoRos required to reconfigure into a sphere. Similarly, the number of limbs is decided by the edges on each face of the platonic solid and the module-topology curve. For example, the cube as the platonic solid and a sinusoidal module-topology curve will result in a four-limb robot where six such MSoRos can reconfigure into a sphere. Similarly, tetrahedron and dodecahedron based MSoRos will result in three and five-limb robot designs that will require four and twelve robots for reconfiguration respectively. Topologically, the spherical configuration of the MSoRo is obtained through inverse orthographic projection of the module-topology curve drawn on the topology curve plane onto the circumscribing sphere. Thereafter, the planar configuration is obtained using azimuthal projection onto a plane tangential to the center of module. The sphere-to-plane distortions are quantified using inter and intramodular distortion metrics. Consequently, the robot topology is optimized by minimizing the difficulty of reconfiguration (distortion) and locomotion. The optimal module-topology curves (sinusoidal family) are calculated for all the five platonic solids. Realizing a functioning robot after this analysis requires consideration of material properties, actuator properties and integration of an actuator into the soft body. The Motor-Tendon Actuator (MTA) limbs of the MSoRos are actuated by motor-tendon drive that are comprised of a motor, spool, tendon and anchor points. The design of the limb cavity is dictated by material properties and the radius of the desired circumscribing sphere. The effect of cavity geometry on the limb stiffness and curling ability is simulated by analyzing a movement of a limb section. The `outward trapezoid' cavity geometry is determined to experimentally have the adequate stiffness and curling ability for the TMA limbs. Four-limb MSoRos are fabricated using casting technique with silicone rubber and the electronics are located in the center of the robot. The supplemental videos show the spherical reconfiguration of six four-limb MSoRos and their planar locomotion. The results both support the design methodology presented in this paper and highlight the exciting potential of MSoRos to transform between dimensions and perform complex tasks and locomotion modes. While this paper focuses on the topology and morphology design and does not include docking, magnetic docking has been successfully implemented in other modular soft robotic systems \cite{kwok_magnetic_2014, zou_reconfigurable_2018}. Future work on this MSoRo system will include docking and multi-modal (i.e., multi-dimensional) locomotion.

\section*{ACKNOWLEDGMENT}
The authors thank Dalton Stacks and Emma Scott for their help. This work is supported by the National Science Foundation under Grant No. 1830432.

\bibliographystyle{IEEEtran}
\bibliography{softrefs}
\end{document}

%% file: 02_Introduction.tex
\begin{figure*}
\begin{center}
\includegraphics[trim = 5pt 10pt 10pt 10pt,clip=true,width=0.75\textwidth]{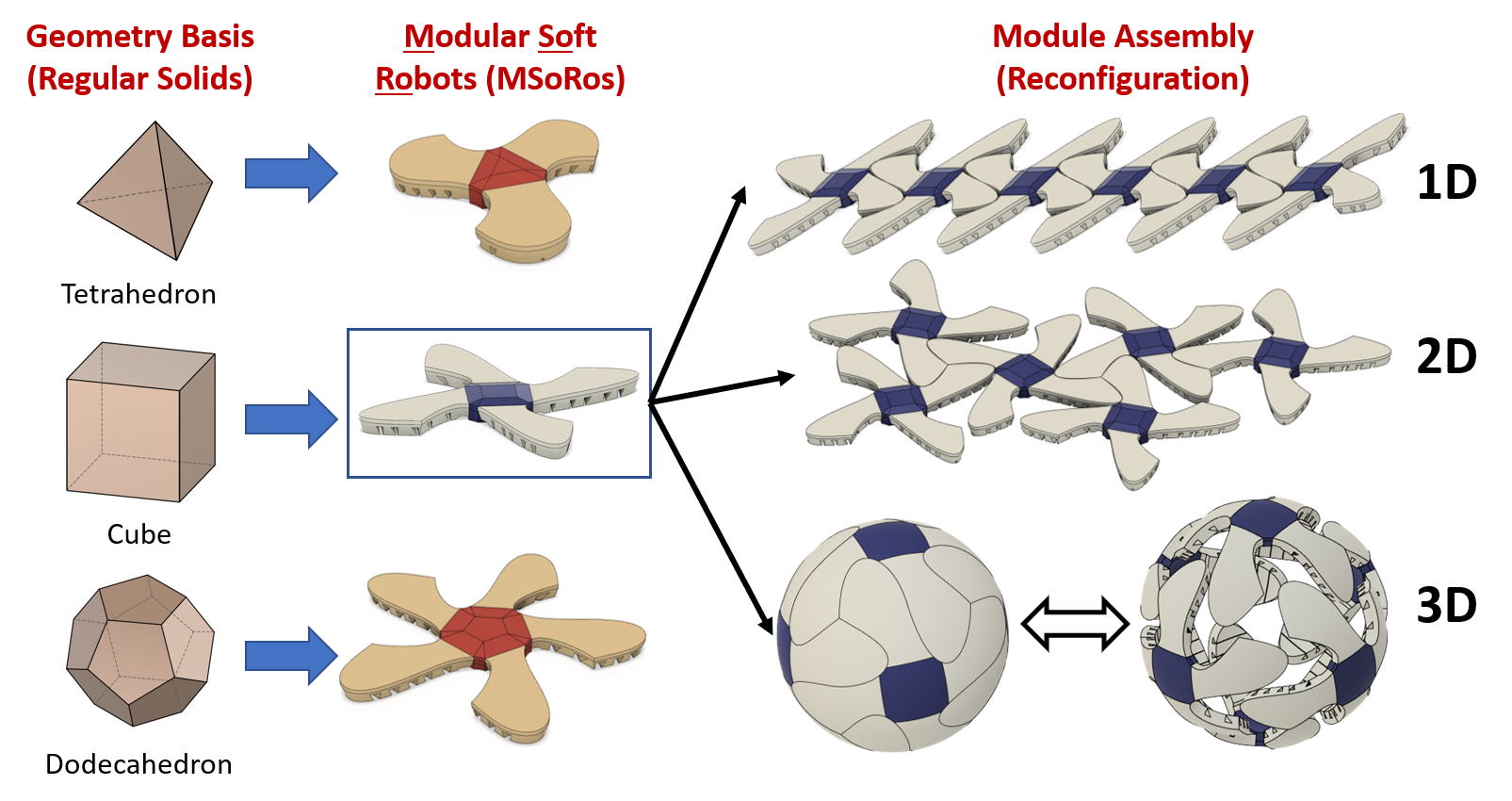}
\caption{Collective morphing by MSoRos between one dimension (caterpillar-like orientation), two (planar swarm) and three (sphere) dimensions. The  three, four and five limb modules result from one of the five platonic solids.}
\label{Fig:Metamorphosis}
\end{center}
\end{figure*}

\section{Introduction} \label{Sec:Introduction}
Biological organisms are characterized by modularity, symmetry and repetitiveness that presumably simplify their development and control of movements. These features also enhance the versatility and robustness of self-reconfigurable robots where the interchangeability of modules (lack of individual identity) enables easy assembly/disassembly for constructing complex robotic systems \cite{yim_modular_2009}. Traditionally, robot modules are rigid and have been shown to carry out different functions including metamorphosis of multiple MTRAN-II modular robots into a four-legged walker, caterpillar or a planar loop \cite{yim_modular_2007, kurokawa_m-tran_2003,lee_dynamic_2002,sastra_dynamic_2009}. In this context, the use of soft materials in robotic modules has the potential to further enhance their versatility, robustness, scalability and customizability.
Unlike traditional rigid robots, soft robots can realize and even transform between complex shapes and curvatures \cite{booth_omniskins_2018}. A swarm of planar (2D) terrestrial, multi-limb Modular Soft Robots (MSoRos) has the potential to display change of dimensions - reconfigure into a spherical ball (2D$\rightarrow$3D) or a caterpillar-like chain (2D$\rightarrow$1D), Fig. \ref{Fig:Metamorphosis}. %
In this case, the multi-module reconfiguration (assembly) has different dynamics as compared to those of individual modules. This can assist the robot to perform tasks including locomotion on variable terrains, e.g., locomotion on uniform hard surface or climbing on a vertical surface may require maximizing the area of contact using a 2D sheet; while on an unstructured terrain, a 3D soccer-ball structure with minimum contact will be most effective for locomotion. Realizing such MSoRos poses design challenges with regard to robot morphology, actuator placement and inter-module docking. The presented research, for the first time, focuses on the topology and morphology design of MSoRos that can exist in dual configurations - planar and spherical. The resulting multi-limb MSoRos are capable of performing terrestrial locomotion on a planar surface while also possessing the geometric ability to morph into a sphere configuration.

Carl Friedrich Gauss' Theorema Egregium, states that the Gaussian curvature of a surface does not change if one bends the surface without stretching it, e.g., a cylindrical tube can be unrolled into a plane as they both have curvature of zero. However, a sphere of radius $R$ having positive curvature of $1/R^{2}$ cannot be flattened into a plane without distorting distances. As an example, in cartography, there is no perfect planar map of Earth and every projection distorts distances. The deformable nature of soft materials makes them appealing as design materials for robots that need to accommodate for the distortions from such transformations. For example, the ability of soft silicone rubber to stretch and shrink, rather than exclusively bend like paper, makes soft materials ideal for designing robots that can compensate for distortions. 
Investigation into tiling of spheres by mathematicians and computer graphics community provide insight for developing techniques to topologically design multi-limb MSoRos \cite{bart_math_nodate,yen_escher_2001}. 
Similarly, the distortion between sphere and plane projections have been quantified using multiple metrics, e.g., Tissot's Indicatrix \cite{snyder_map_1987}. The MSoRos, if designed appropriately, can potentially conform seamlessly between planar and spherical configurations where the resulting distortions are mitigated by the soft material properties. The associated challenges can be categorized as topology design of spherical and planar configurations, and the morphology design for physical realization of a robot that facilitates smooth transition between the configurations.
Mechanically, soft robot are traditionally actuated using hydraulic and pneumatic fluid actuators, electroactive polymers, and variable length tendons \cite{rus_design_2015}. Motor-Tendon Actuation (MTA) provides advantages of precision control, high bandwidth and light weight. However, the integration of these actuators into soft body of a terrestrial soft robot is challenging. Unlike rigid robots, the placement of MTA inside the robot body directly influences the deformation profile of the robot and its locomotion behavior \cite{vikas_design_2016}.

\textit{Contributions:} The research presents a design approach for topology and morphology design of homogenous MSoRo that can exist in dual configurations - planar and spherical. (1) \textit{Geometric basis of robot topology design}. The design of these topologies is based in geometry that involves choice of one of the five platonic solids and a module-topology curve. The type of polyhedron characterizes the multi-module assembly - the number of modules required for reconfiguration and radius of the reconfigured sphere correspond to the number of polyhedron faces and radius of the circumscribing sphere respectively. The module-topology curve is an odd-function (to ensure homogeneity) drawn about the polyhedron edge along the plane with normal vector joining the center of the polyhedron and the edge. This curve determines the robot topology in the different configurations. (2) \textit{Scale-invariant topology design through projections}. Spherical topology is obtained by inverse orthographically projecting the module-topology curves about each of the polygon edges onto the circumscribing sphere. The azimuthal projection of the spherical topology onto the tangent plane about the point joining the center of the sphere and the polyhedron face results in the planar topology of the robot. The scaling of base platonic solid is reflected linearly in spherical and planar topologies. %
(3) \textit{Intermodular and intramodular distortion metric to quantify for reconfiguration difficulty.} The intermodular distortion is defined as the least possible overlapping surface area between two adjacent modules in planar configuration to the area normalized w.r.t. the module surface area. The intramodular distortion correlates to the material properties. These metrics assist in formulating the cost function that optimizes the difficulty of locomotion with that of reconfiguration to determine the optimal module-topology curve. (4) \textit{Adjusting robot limb cavity geometry for tuning its stiffness and curling ability}. Fabrication of a finite-width MTA MSoRos capable of smooth transition between zero and positive Gaussian curvature involves manipulation of limb stiffness and curling ability. This is achieved by introducing cavities of different geometries without compromising the structural integrity of the limb actuator. 

The research illustrates design of three, four or five limbed MSoRos (based on tetrahedron, cube and dodecahedron respectively) that can reconfigure into a sphere using four, six or twelve modules. The paper is structured as follows: Sec \ref{Sec:Topology} describes the scale invariant topological design of MSoRos in spherical and planar configurations. Optimization for the module-topology curve that minimizes intermodular distortion (between neighboring modules) and locomotion difficulty is presented in Sec \ref{Sec:Distrotions}. Thereafter, the manipulation of MTA limb stiffness and curling ability by introducing cavities of different geometries is described in Sec \ref{Sec:Cavities}. Finally, Sec \ref{Sec:Fabrication} discusses fabrication a four limb MSoRo capable of planar locomotion and spherical reconfiguration. 

%% file: 03_TopologyDesign.tex
\section{Scale-invariant Spherical and Planar Topology Design} \label{Sec:Topology}

\begin{figure*}
\begin{center}
\includegraphics[width=0.75\textwidth]{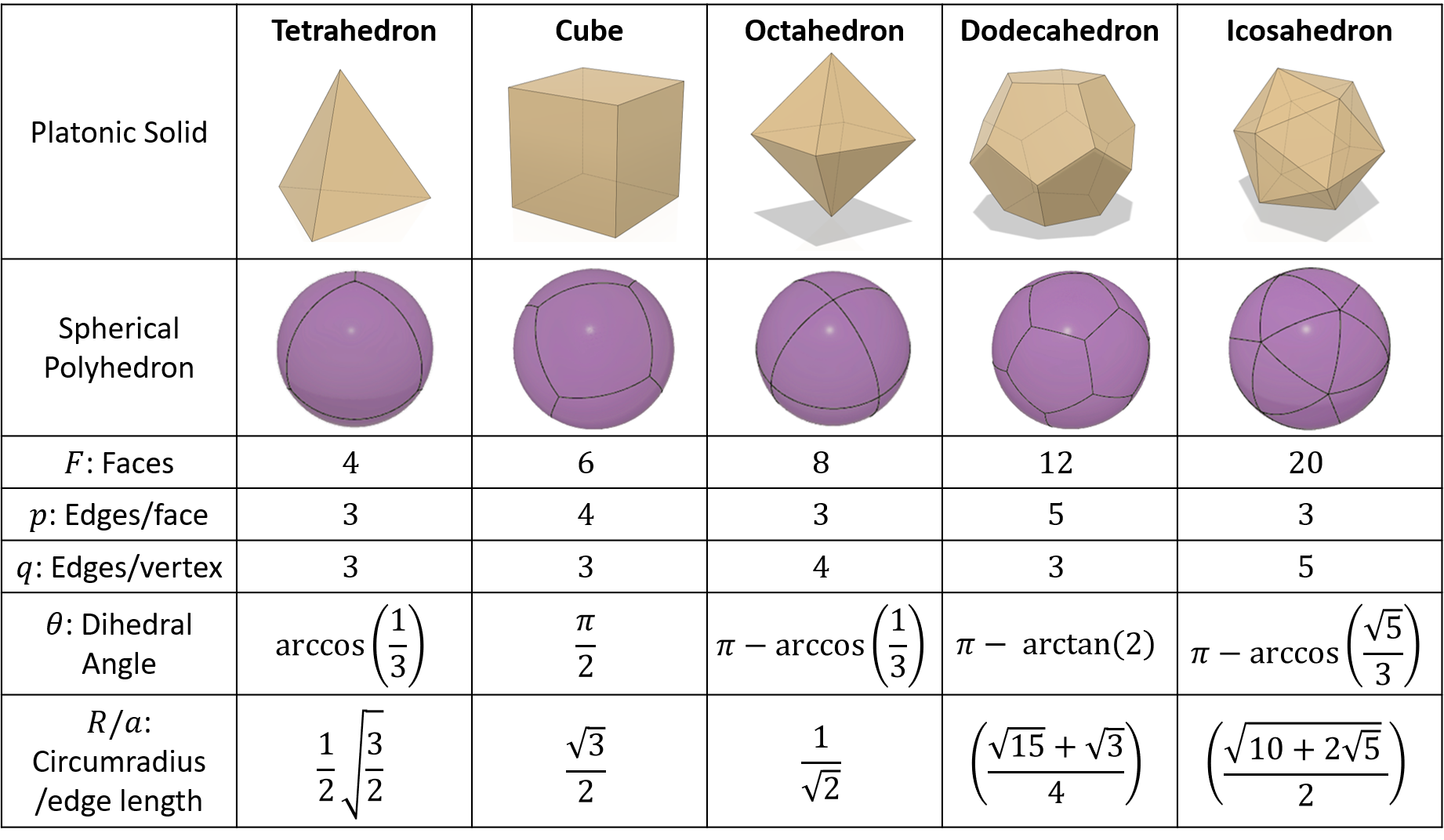}
\caption{Projection of all the five platonic solids of edge length $a$ onto a circumscribing sphere of radius $R$. The number of faces $F$, edges per face $q$, and the dihedral angle $\theta$ correlate to the number of modules required for spherical reconfiguration, the number of limbs per module and the topology curve plane respectively.}
\label{Fig:PlatonicSolidsSpherical}
\end{center}
\end{figure*}

Challenges associated with topology design of planar robots capable of spherical reconfiguration can be identified as controlled tessellation of a sphere and their subsequent projection onto a plane. With regard to the former, mathematicians Delp and Thurston \cite{delp_playing_2011} have formulated a process of using complex sketches on octahedral faces to approximate spherical tiling with paper, a specimen with relatively high tensile stiffness in the plane. Conceptually, they attempt to find two curves where a neighborhood of the curve on the surface can be matched to a neighborhood of the curve in the Euclidean plane. Similarly, researchers have studied how origami tessellations can be optimized to approximate surfaces of varying curvature \cite{dudte_programming_2016}. For the latter challenge relating to `unwrapping' of a sphere onto a plane, a conformal projection preserving shape would be ideal. However, such a projection that is conformal over large range of angles is not possible. More precisely, a distortion-free map between surfaces of zero (e.g., plane) and non-zero (e.g., a sphere) Gaussian curvature does not exist. The resulting distortions manifest as inaccuracies in distances, angles (shapes), areas or their combinations. Much of the literature on this type of mapping, especially in the fields of cartography and geography, focuses on constructing projections which minimize such distortions \cite{snyder_map_1987}.  The ability of soft materials (e.g., silicone rubber) to stretch and shrink, rather than exclusively bend like paper, makes them ideal for designing robots that can reconfigure between a plane and a sphere.

Tessellation of a sphere refers to tiling of a sphere without any overlaps and gaps. Spherical polyhedrons provide natural inspiration for tesselating a sphere with identically shaped tiles. They are spherical equivalent of platonic solids where the projections preserve solid angles and vertices (point of intersection between the polyhedron and the circumscribing sphere). This projection of the polyhedron onto the sphere does not preserve the circumference, the surface area or the volume. In three-dimensional space, polyhedrons are solids that are composed of flat surfaces with straight edges and sharp corners (vertices). Platonic solids are convex regular polyhedron having regular faces and solid angles (vertex figures), i.e., all faces are same regular polygon with identical sides and angles \cite{coxeter_regular_1973}.  Only five solids meet this criterion - tetrahedron (four faces), cube (six faces), octahedron (eight faces), dodecahedron (twelve faces) and icosahedron (twenty faces) \cite{coxeter_regular_1973}.  The relationship between the edge length of the platonic solid $a$, the dihedral angle $\theta$ (the angle between the adjescent faces) and the circumradius $R$ can be summarized as \cite{zwillinger_crc_2002}
\begin{align}
\theta = 2 \mathrm{arcsin}\left(\frac{\cos\left(\pi/q\right)}{\sin\left(\pi/p\right)}\right)\nonumber\\
\frac{R}{a} = \frac{\sin\left(\pi/q\right)}{2\sin\left(\pi/p\right)}\sec\left(\frac{\theta}{2}\right)
\end{align}
where $p,q$ are the number of edges in a face and number of edges meeting at a vertex respectively, Fig. \ref{Fig:PlatonicSolidsSpherical}.

\begin{figure}[h]
\begin{center}
\includegraphics[width=\columnwidth,trim={5pt 5pt 5pt 5pt},clip]{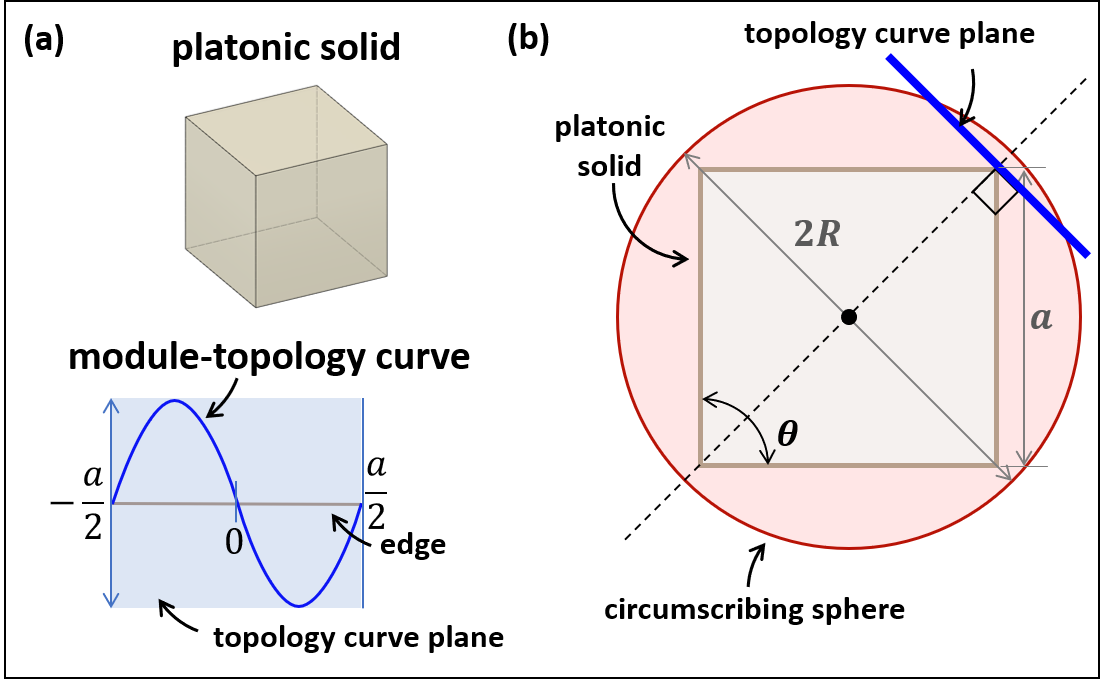}
\caption{The design process is based in geometry where (a) the designer has choice of a platonic solid (cube as a visual example) and module-topology curve. (b) The module-topology curve is drawn along the polyhedron edge on a topology curve plane (blue). The normal to the plane (dotted line) is the vector joining the centers of the circumscribing sphere (red) and the edge of the platonic solid. The dihedral angle $\theta$ is the angle between adjacent faces.}
\label{Fig:GeometricBasis}
\end{center}
\end{figure}

Platonic solids are central to design spherical and planar topologies of rotationally symmetric, homogeneous MSoRos. More complex module combinations may be derived using convex regular-faced polyhedron as the base solid. This alteration of the base solid is capable of producing rotationally symmetric heterogeneous and asymmetric homogeneous modules. However, we focus our discussion to rotationally symmetric, homogeneous MSoRos where the designer has choice of \textit{the base platonic solid}, and \textit{the module-topology curve}. They determine the characteristics of the multi-module assembly. The number of faces of the platonic solid determine the number of modules that will reconfigure to form a sphere. For example, four and six homogeneous MSoRos based on a tedrahedron and cube respectively, will be required for spherical reconfiguration. The module topology curve is drawn on a plane that is normal to the line joining the centers of the circumscribing sphere and the edge of the platonic solid, Fig. \ref{Fig:GeometricBasis}. This plane passes through the edge and is equally inclined at $\displaystyle \left(90^o - \frac{\theta}{2}\right) $ from the face of the polyhedron where $\theta$, the dihedral angle, is the angle between adjacent faces.

The mathematical equivalent of designing homogenous, rotationally symmetric modules is isohedral tiling. Simply put, isohedral spherical tiling implies covering a sphere with identical modules without any gaps or overlaps, only by performing translations and rotations on the single tile. The constraint of rotational symmetry implies that the module-topology curve drawn along each edge and face is the same. Let this curve be denoted by $f(x)$ s.t. $x\in[-\frac{a}{2},\frac{a}{2}]$. Imagine adjacent faces such that the curves corresponding to each module along the common edge are $f(x)$ and $f(y)$ respectively. The condition of no gaps or overlaps is only satisfied when the curve is invariant under $180\deg$ rotation about the center of the edge while being continuous at the edges. Equivalently, constraining the module-topology curve to be an odd-function with constraints at the edges
\begin{align}
\begin{bmatrix}
x \\
f(x)
\end{bmatrix} &= 
\begin{bmatrix}
\cos \pi & -\sin \pi \\
\sin \pi & \cos \pi
\end{bmatrix}
\begin{bmatrix}
y \\
f(y)
\end{bmatrix} \nonumber\\
f(-x) &= -f(x) 
\qquad x\in\left[-\frac{a}{2},\frac{a}{2}\right] \label{Eqn:Curve180Rotation}\\
s.t. \quad f\left(\frac{a}{2}\right) &= f\left(-\frac{a}{2}\right)=0 \nonumber
\end{align}
This provides the designer with multitude of choices. However, for the remainder of the discussion we consider the sinusoidal family of functions 
\begin{align}
    f(x) = A\frac{a}{2}\sin\left(
    \frac{2\pi}{a}x\right) \label{Eqn:CurveFamily}
\end{align}
where $\displaystyle A\in\left[-1,1\right]$. For this family of curves with one maxima and minima, the number of limbs per module is equal to the number of edges $E$ of the face polygon.

\begin{figure}[h]
\begin{center}
\includegraphics[width=\columnwidth,trim={5pt 10pt 10pt 5pt},clip]{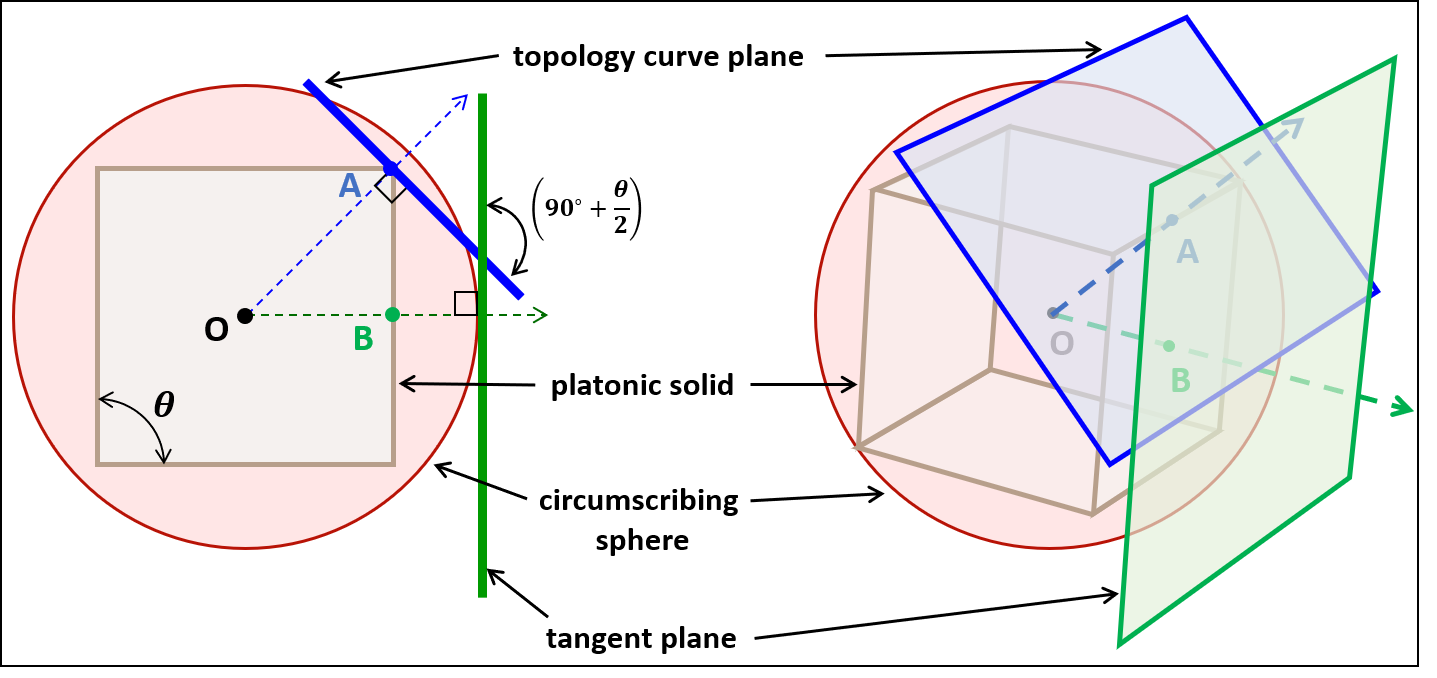}
\caption{The topology curve plane (blue) and the tangent plane (green) for cube as the base platonic solid. The topology curve plane passes through the edge of the polyhedron with the normal vector along the line joining the centers of the circumscribing sphere (red) $O$ and the edge $A$. The tangent plane normal is along the vector joining $O$ and the center of the polyhedron face $B$.}
\label{Fig:PlanesVisualization}
\end{center}
\end{figure}

\begin{figure}
\begin{center}
\includegraphics[width=\columnwidth,trim={7.5pt 12.5pt 20pt 1pt},clip]{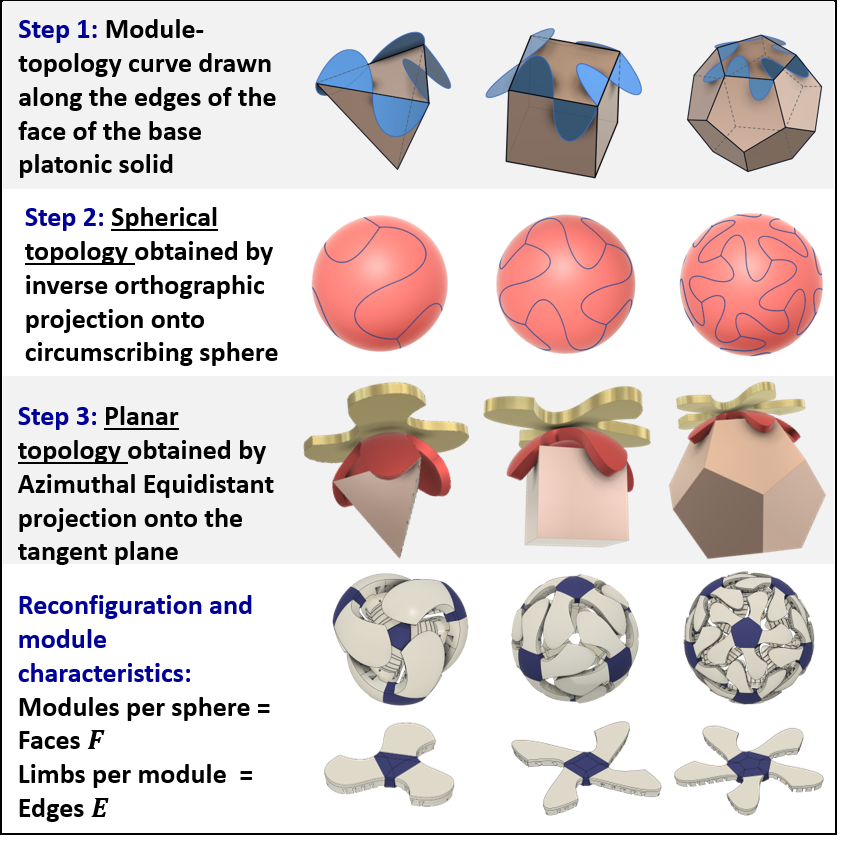}
\caption{Summary of the sequential methodology for designing homogeneous, rotationally symmetric multi-limb MSoRos. The designer has choice of base platonic solid and the module-topology curve. The outcome is spherical and planar configurations of MSoRos.}
\label{Algorithm}
\end{center}
\end{figure}

\textit{Spherical Topology}. The spherical topology is obtained by inverse orthographic projection, $\bm{g_0}:\mathbb{R}^2 \rightarrow S^2$, of the module topology curve $f(x)$ onto the circumscribing sphere. Given radius of the sphere $R$, and $(\phi_0,\lambda_0)$ as the latitude and longitude of the origin of the projection, the latitude and longitude of the projection $(\phi_s,\lambda_s)$ are \cite{snyder_map_1987}
\begin{align}
\phi_s &= \arcsin\left[\cos c \sin \phi_0 +\frac{f(x) \sin c \cos \phi_0}{\rho} \right] \nonumber \\
\lambda_s & = \lambda_0 + \arctan\left(\frac{x \sin c}{\rho \cos c \cos \phi_0 - f(x) \sin c \sin \phi_0} \right) \label{Eqn:ProjectedSphere}
\end{align}
\begin{align}
\begin{bmatrix}
\phi_s\\
\lambda_s
\end{bmatrix} &= %
\begin{bmatrix}
\arcsin\left[\cos c \sin \phi_0 +\frac{f(x) \sin c \cos \phi_0}{\rho} \right] \nonumber\\
\lambda_0 + \arctan\left(\frac{x \sin c}{\rho \cos c \cos \phi_0 - f(x) \sin c \sin \phi_0} \right) 
\end{bmatrix} \\&= \bm{g_0}(x,f(x),\phi_0,\lambda_0)
\end{align}
where $\rho = \sqrt{x^2 + f(x)^2}, c =\arcsin\left({\rho}/{R}\right)$. For the example edge shown in Fig. \ref{Fig:PlanesVisualization} with the base solid as a cube, $\lambda_0=45^{\circ},\phi_0 =0$. Most CAD softwares are capable of such orthographic projection and repetition of this process about each edge results in monohedral tiling of the sphere, Fig. \ref{Algorithm}.

\textit{Planar Topology}. The projection of the spherical tile onto a planar surface is necessary for robot fabrication, especially, with finite width that can allow actuators to be incorporated. The azimuthal equidistant projection is chosen as it preserves distance and direction from the center of the projection. Hence, minimal distortion occurs close to the center, where, as it will be evident later, the least flexible materials (e.g. rigid-flexible motor-tendon actuators) will be situated.  The deformability of the soft material (silicone rubber in this case) accounts for the area distortions as we move away from the center. Consequently, as illustrated in Fig. \ref{Fig:PlanesVisualization}, the plane of projection is the tangent plane (green) to the circumscribing sphere (red) with the normal passing through the centers of the sphere and the polyhedron face (dotted line from $O$ to $B$). Geometrically, the rectangular coordinates for Azimuthal Equidistant projection,  $\bm{g_1}:{S}^2 \rightarrow \mathbb{R}^2$, given sphere radius $R$ and center of projection $(\phi_1,\lambda_1)$ are
\begin{align}
\begin{bmatrix}
x_p\\
y_p
\end{bmatrix} &=
Rk'\begin{bmatrix}
 \cos \phi \sin \left(\lambda -\lambda_1\right) \nonumber\\
 \cos \phi_1 \sin \phi -\sin \phi_1 \cos \phi \cos (\lambda -\lambda_1)
\end{bmatrix}\\
&= \bm{g_1}\left(\bm{g_0}(\phi_0,\lambda_0),\phi_1,\lambda_1\right)\\
&= \bm{h}\left(x,f(x),\phi_0,\lambda_0,\phi_1,\lambda_1\right)\nonumber
\label{Eqn:ProjectedPlane}
\end{align}
where $\displaystyle k'=\left({c}/ {\sin c}\right)$ is the scale factor, and $\cos c = \sin \phi _1 \sin \phi + \cos \phi_1 \cos\phi \cos(\lambda-\lambda_1)$. For the example tangent plane in Fig. \ref{Fig:PlanesVisualization}, $\phi_1=\lambda_1=0$.
Scale invariance refers to characteristic of the system's structural properties to remain unchanged (invariant) at different scales. For our case, the planar and spherical topologies are scale invariant - the linear scaling of the planar topology implies linear scaling of the reconfiguring sphere. Practically, from the perspective of a designer, fabrication of a linearly scaled MSoRo will result in scaling of the reconfigured sphere by the same factor.\vspace{5pt}\\
\textit{Proposition:} For a linearly scaling modular topology curve, the scaling of the base platonic solid is reflected linearly in the spherical and planar topologies.
\begin{align}
\bm{h}(\mu x, f(\mu x)) = \mu \bm{h}(x, f(x)), \nonumber\\ \bm{g_0}(\mu x, f(\mu x)) = \bm{g_0}(x, f(x)) \\
\forall f(\mu x)  = \mu  f(x)~\mathrm{s.t.}~ \mu \in \mathbb{R} \nonumber
\end{align}
\textit{Proof:}  Linear scaling $\mu x$ implies scaling of the base platonic solid edge and the circumscribing sphere radius by the same factor, i.e., $a(\mu x) = \mu a(x), R(\mu a) = \mu R(a)$. Considering the spherical topology,
\begin{align*}
\rho(\mu x) = \sqrt{(\mu x)^2 + f(\mu x)^2} = \mu \rho(x), \quad \frac{f(\mu x)}{\rho (\mu x)} = \frac{f(x)}{\rho (x)}\\
c(\mu x) = \mathrm{arcsin}\left(\rho(\mu x)/R(\mu x)\right) = c(x)\\
\Rightarrow \bm{g_0}(\mu x, f(\mu x)) = \bm{g_0}(x, f(x))
\end{align*}
Similarly, for the planar topology, 
\begin{align*}
k'(\mu x) = k'(x) 
\Rightarrow \bm{h}(\mu x, f(\mu x)) = \mu \bm{h}(x, f(x)) \qquad \blacksquare
\end{align*}

The regular faces of the five platonic solids are either a triangle, square or a pentagon (Fig. \ref{Fig:PlatonicSolidsSpherical}). Consequently, tetrahedron, cube and dodecahedron will be considered as base platonic solids for designing three, four and five limb robots. The methodology can easily be extended to design homogenous, symmetrical three limb MSoRos using octahedron and icosahedron. Here, eight and twenty modules, respectively, will be required to reconfigure into a sphere. The Fig. \ref{Algorithm} sequentially articulates the methodology of MSoRo design from design choice of platonic solid and the module-topology curve for the three polyhedrons.

%% file: 04_DistortionQuantification.tex
\section{Optimal Module Topology for Locomotion and Reconfiguration Ability} \label{Sec:Distrotions}
\begin{figure}
\begin{center}
\includegraphics[width=\columnwidth,trim={15pt 5pt 10pt 10pt},clip]{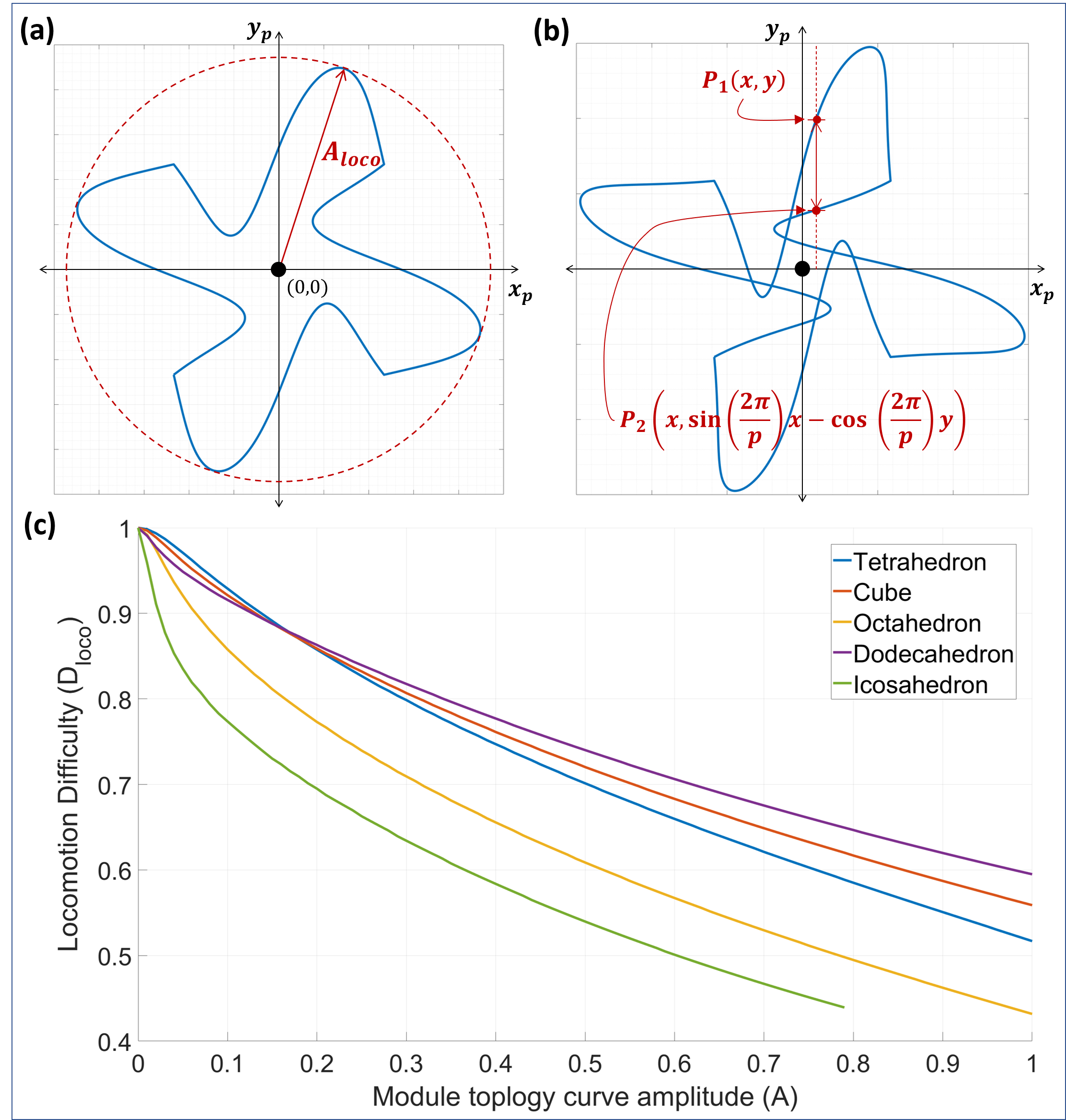}
\caption{For an sinusoidal curve amplitude $A$, (a) the limb length $A_{loco}$ is proportional to ease of locomotion. (b) The existence of the planar topology is ensured only when the curves do not intersect and the distance between points $P_1$ and $P_2$ is more than $c_{slack}$. For example, this topology is unfeasible as the curves intersect. (c) The difficulty of locomotion as a function of $A$ for all the platonic solids increases as the amplitude increases, i.e., `longer' limb. The metric $D_{loco}$ does not exist for $A>0.79$ with the icosahedron as the base platonic solid as the topology is infeasible.}
\label{Fig:Aoptimum}
\end{center}
\end{figure}

The topology curve has influence on the locomotion and reconfiguration abilities of the MSoRos. Conceptually, locomotion is result of optimizing forces at different parts of the body \cite{radhakrishnan_locomotion:_1998}. For a soft module, the length of the limb enhances its ability to interact with the environment. We introduce the concept of \textit{locomotion ability} that quantifies the versatility of the robot module for a variety of locomotion modes under unknown environmental conditions, e.g.,  maneuvering obstacles using planar and spherical configurations in 1D, 2D or 3D mode. Here, the locomotion ability is assumed to be proportional to the limb length $A_{loco}$, Fig. \ref{Fig:Aoptimum}a. We define the locomotion difficulty as the inverse of the normalized $A_{loco}$
\begin{align}
\begin{gathered}
A_{loco} = \max \left(x_p^2 + y_p^2\right) \\
s.t.~\left|y_p - \sin\left({2\pi}/{p}\right)x_p-\cos\left({2\pi}/{p}\right)y_p\right|>c_{slack} \\
 x_p\in \left[-{b}/{2},{b}/{2}\right]
\end{gathered}
\end{align}
where the curve amplitude is constrained such that the planar topology is feasible, i.e., the distance between the curve and the rotated curve, Fig. 6b, is always greater than a specified minimum $c_{slack}$. We define the difficulty of locomotion $D_{loco}$ as the inverse of the normalized limb length
\begin{align}
D_{loco}(A) &= \frac{\left(\frac{1}{A_{loco}(A)}\right)}{\max_{A\in[0,1]}\left(\frac{1}{A_{loco}(A)}\right)}
\end{align}
The Fig. \ref{Fig:Aoptimum}c plots the $D_{loco}$ for the five platonic solids as the module topology amplitude is varied. Due to infeasible planar topology, $D_{loco}$ and $A_{loco}$ are not defined for $A>0.79$ with icosahedron as the based platonic solid.

The metrics of quantifying distortions provide insight about the capability of the robot module to reconfigure. Multiple distortion metrics have been proposed to quantify sphere-plane distortions, e.g., Tissot's Indicatrix \cite{snyder_map_1987}. The reconfiguration ability of MSoRos compels us to quantify distortions between sphere-and-plane configurations of a single module and those between two adjacent modules (in planar configuration). We refer to these dimensionless parameters as intramodular and intermodular distortions respectively.
\begin{figure}
\begin{center}
\includegraphics[width=0.95\columnwidth,trim={15pt 2pt 2pt 10pt},clip]{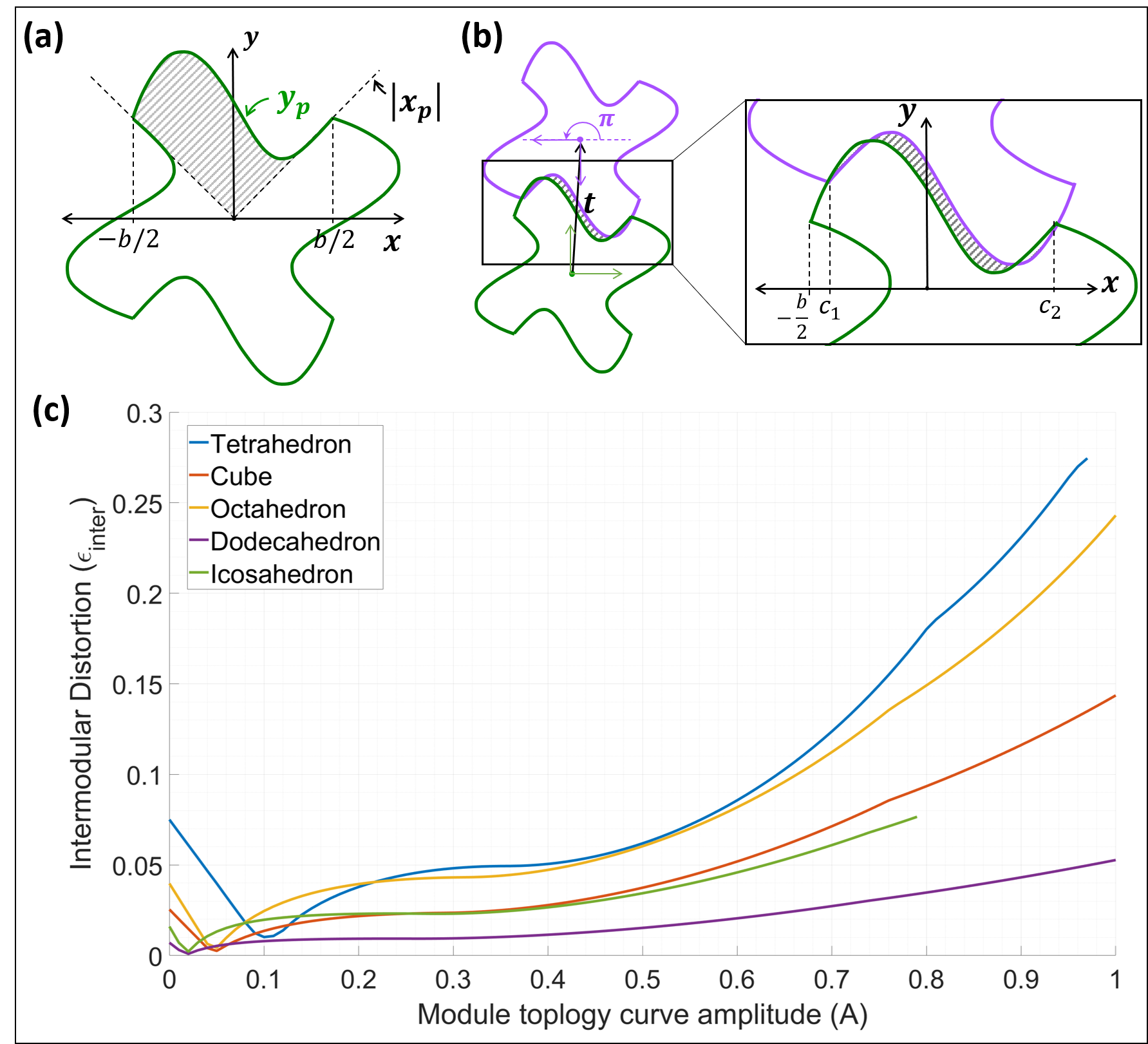}
\caption{For a tetrahedron-based four limb MSoRo: (a) The intramodular distortion is defined by the ratio of module surface area in planar and spherical configuration. The limb sector area (shaded) is the area between the projected module topology curve $y_p$ and the limb sector line $|x_p|$. (b) The intermodular area $G_E$ is defined as the minimum area of overlap (shaded) between adjacent modules separated by displacement $\bm{t}$ in planar configuration. (c) For all the platonics solids, the intermodular distortion increases after the initial decrease as $A$ increases. This trend is near-opposite to the one observed for the module locomotion difficulty.}
\label{Fig:Distortions}
\end{center}
\end{figure}
\begin{figure*}
\begin{center}
\includegraphics[width=0.9\textwidth,trim={50pt 50pt 50pt 10pt},clip]{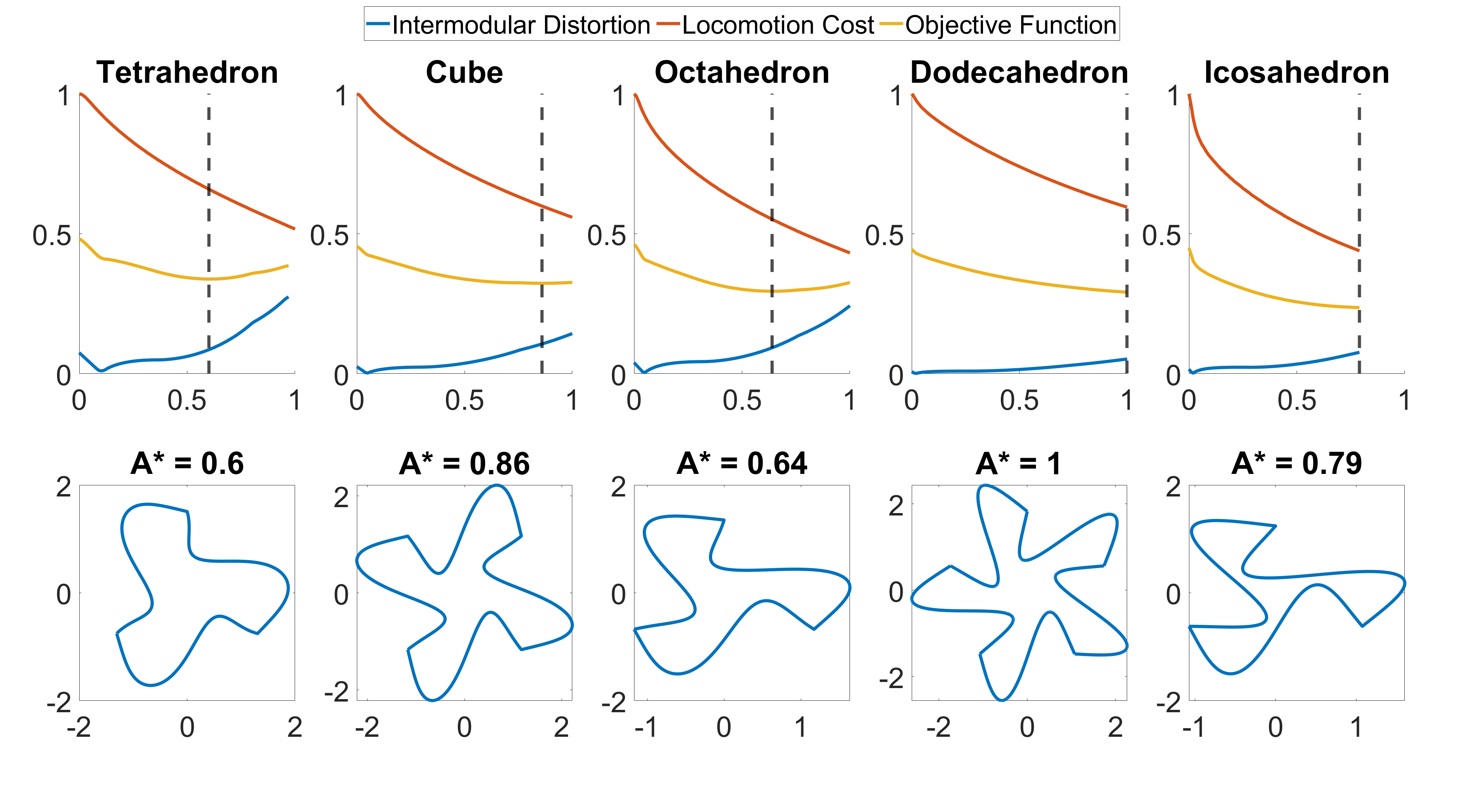}
\caption{Planar topologies with the five base platonic solids optimizing for locomotion and reconfiguration ability with weight $\alpha=0.56$. The optimal values ($A^*$) of the three limb modules resulting from tetradehron, octahedron and icosahedron as the base platonic solids increases as the number of modules required for spherical reconfiguration also increases from 4, 8 to 20.}
\label{Fig:OptimalPlanarTopologies}
\end{center}
\end{figure*}
{Intramodular distortion}, $\epsilon_{intra}$, is the ratio of the surface areas of the top of a single MSoRo module in planar $A_E$ and spherical configurations $A_S$. The former is calculated as the number of edges multiplied by the limb sector area - the area between the curve of the projected module-topology curve $y_p$ and the limb sector line $\displaystyle \left|\cot\left(\frac{\pi}{p}\right) x_p\right|$, Fig. \ref{Fig:Distortions}a. The latter can be obtained by observing that the homogenous spherical monohedral tiling divides the area of the sphere equally among each of the modules. 
\begin{align}
\begin{gathered}
\epsilon_{intra} = \frac{A_{E}-A_{S}}{A_{E}} \quad\mathrm{where} \quad A_S = \frac{4\pi R^2}{n}\\
A_E = n\int_{-\frac{b}{2}}^{\frac{b}{2}}\left(y_p -\left|\cot\left(\frac{\pi}{p}\right) x_p\right|\right)dx_p
\end{gathered}
\end{align}
where $[x_p,y_p]^T=\bm{h}\left(x,f(x),\phi_0,\lambda_0,\phi_1,\lambda_1\right)$  and $\frac{b}{2} = x_p\left(\frac{a}{2},0,\phi_0,\lambda_0,\phi_1,\lambda_1\right)$ for a given edge and face (Eqn. \ref{Eqn:ProjectedPlane}). This parameter quantifies the necessary surface deformation required for transitioning between planar and spherical configurations. This metric can be used as a design guide to optimize for factors directly affecting the surface deformation, e.g., actuation, material and morphology selection.

The intermodular area $G_E$ is defined as the least possible overlapping surface area between two adjacent modules in planar configuration. Unlike for spherical configuration, the non-zero area is dependent on the distance between the centers of the two modules $\bm{t}=[t_1,t_2]^T$, Fig. \ref{Fig:Distortions}b. Considering the rotational symmetry and assuming that the modules are placed such that the module curves are rotated by $180 \deg$, the rotated curve $(x_p',y_p')$  and the infinitesimal area between the curves $dA$ is
\begin{align*}
\begin{bmatrix}
x_p'\\y_p'
\end{bmatrix}&= 
\begin{bmatrix}
t_1\\t_2
\end{bmatrix} + %
\begin{bmatrix}
\cos(\pi) &-\sin(\pi)\\\sin(\pi)&\cos(\pi)
\end{bmatrix}
\begin{bmatrix}
x_p\\y_p
\end{bmatrix} \\
dA &= y_p(x_p)dx_p - y_p'(x_p)dx_p' \nonumber\\
&= \left(y_p(x_p) +y_p(-x_p+t_1)-t_2 \right)dx_p
\end{align*}
and the intermodular area is calculated as
\begin{align}
G_E 
&= \min_{\bm{t}}\int_{c_1}^{c_2}\left(y_p(x_p) +y_p(-x_p+t_1) - t_2\right)^2 dx_p 
\end{align}
\newcommand{\range}[2]{\left[{#1}, {#2}\right]}
where the interval $ \range{c_1}{c_2} = \range{-\frac{b}{2}+t_1}{ \frac{b}{2}+t_1} \cap \range{-\frac{b}{2}}{\frac{b}{2}}$, Fig. \ref{Fig:Distortions}b. Subsequently, the intermodular distortion $\epsilon_{inter}$ is defined as 
\begin{align}
\epsilon_{inter}(A) = \frac{G_E}{A_E}
\end{align}

The objective function is defined as the weighted sum of locomotion difficulty cost and intermodular distortion (reconfiguration difficulty). Consequently, the optimal curve amplitude $A^*$ is obtained by minimizing this objective function
\begin{align}
\begin{gathered}
J(A) =  \alpha \epsilon_{inter} + (1-\alpha) D_{loco} , \quad \alpha \in [0,1]\\
A^* = \min_{A} J(A)
\end{gathered}
\end{align}

We numerically optimize the objective functions in MATLAB\textregistered~ for weight $\alpha=0.56$ to obtain $A^*$ and corresponding planar module configuration for the five platonic solids, Fig \ref{Fig:OptimalPlanarTopologies}. The three limb modules resulting from tetradedron, octahedron and icosahedron provide more insight. As the number of modules required for spherical reconfiguration increases, the amount of required distortion also decreases, hence, allowing for longer module limb, i.e., $A^*$.

%% file: 05_Fabrication.tex
\section{Optimizing Cavity Geometry for Limb Stiffness and Curling Ability} \label{Sec:Cavities}
\begin{figure}
\begin{center}
\includegraphics[trim={10pt 5pt 5pt 5pt},clip,width=\columnwidth]{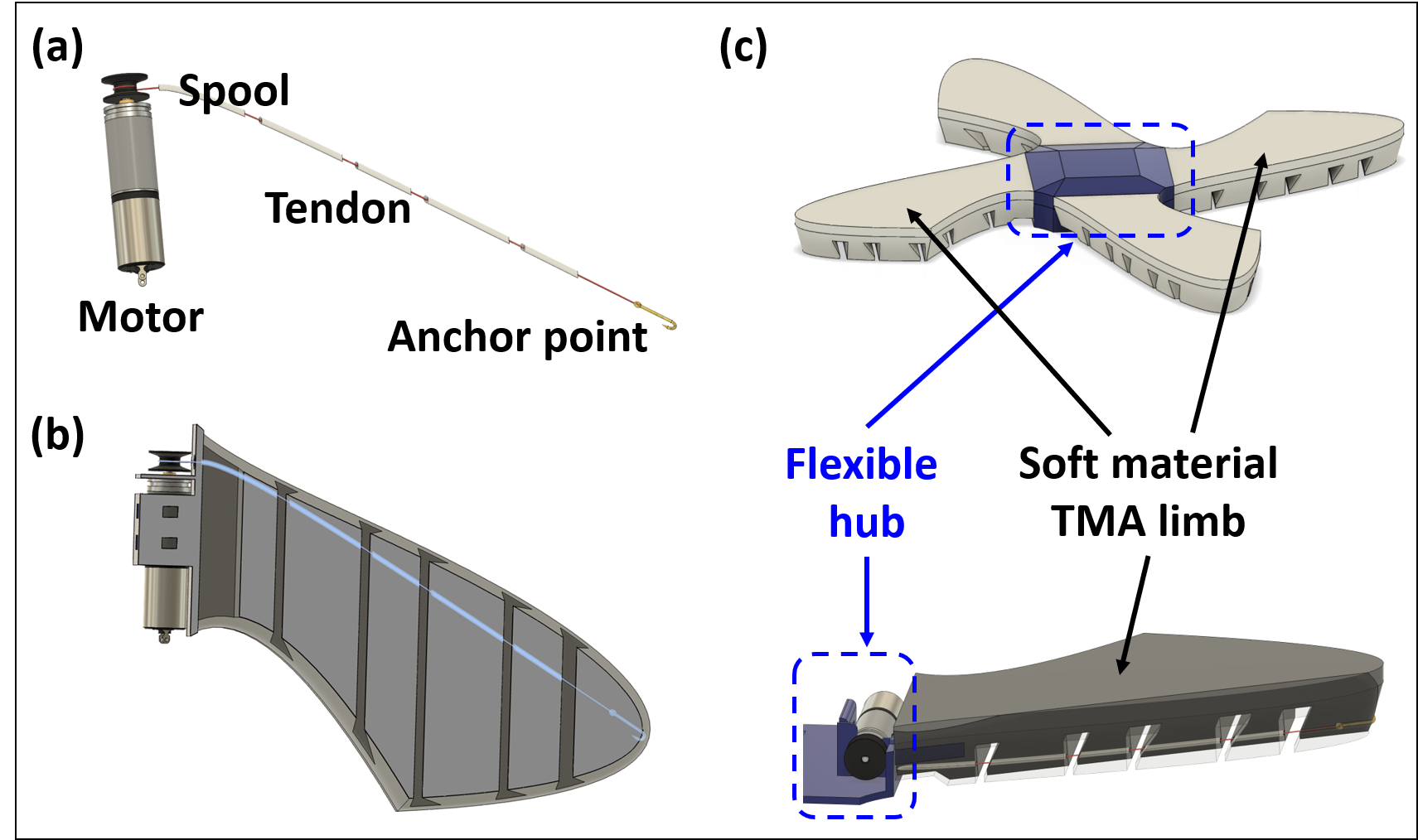}
\caption{(a) Motor Tendon Actuator (MTA) and its components. (b) MTA limb with the actuator embedded inside the soft body. (c) A 4-limb MSoRo comprising of 4 actuators merged at the flexible hub.}
\label{Fig:TMALimb}
\end{center}
\end{figure}

Terrestrial soft robots have been actuated by pneumatic actuators, shape memory alloys,
dielectric elastomers and motor-tendon combination \cite{shepherd_multigait_2011, umedachi_softworms:_2016}. Electro-mechanical actuators have advantage of being more precise, high bandwidth and less electronic payload. Motor Tendon Actuators (MTAs), also known as cable motor actuators, have shown promise in soft robotics, manipulators, and wearable exo-suits applications \cite{laschi_soft_2016,wehner_lightweight_2013}. They use an integrated tendon motor drive to elastically deform the actuator's soft body. By embedding them inside the soft body, the material properties can be manipulated to design lighter, robust, and efficient soft actuators. Additionally, tendon drives are uniquely suited for soft robotics application because of their simplicity, flexibility, and ability to transmit the motor torque as a directional force to remote locations. These attributes result from the four fundamental components of motor, spool, tendon, and anchor points, Fig. \ref{Fig:TMALimb}a. The motor simultaneously supplies the actuation torque, and acts as one of the required anchor points. The spool interfaces the motor and tendon by converting the rotational into linear motion to regulate the speed, tension, and alignment of the attached tendon. The tendon facilitates transmission of force across the body to other (or final) anchor points. The anchor point refer to location or points where the tendon is fixed to the body, component, or any other arbitrarily assigned fixture. The MTA incorporated inside the MSoRo limb is referred to as a MTA limb, Fig. \ref{Fig:TMALimb}b. A flexible material central hub holds the control payload and motors for MTA limbs, Fig. \ref{Fig:TMALimb}c.

The morphology of the limb plays a critical role in design of a finite-width MSoRo that can actuate between configurations with zero and positive curvatures. The topology designs obtained thus far are insightful, however, do not considered complexities of fabrication, e.g., material properties, weight of the electronic payload and actuation constraints. For this discussion, we refer to curling and uncurling as the motions that transition the limb from planar to spherical configuration and vice versa. For MSoRos, a MTA limb must be soft enough to curl to a desired Gaussian curvature and simultaneously lift the weight of the potential payload. At the same time, it should have enough elastic energy in the spherical configuration to uncurl back to the planar configuration without any actuation. We refer to these traits as the limb '\textit{stiffness}' and `\textit{curling ability}'. As a designer, controlled curling can be geometrically achieved by introducing cavities along the thickness of the MTA limb, Fig. \ref{Fig:TMALimb},broadly referred to as morphological computation. Here the manipulation of stiffness and curling ability can be achieved by altering the cavity geometry. 

\begin{figure}
\begin{center}
\includegraphics[trim={5pt 5pt 5pt 5pt},clip,width=\columnwidth]{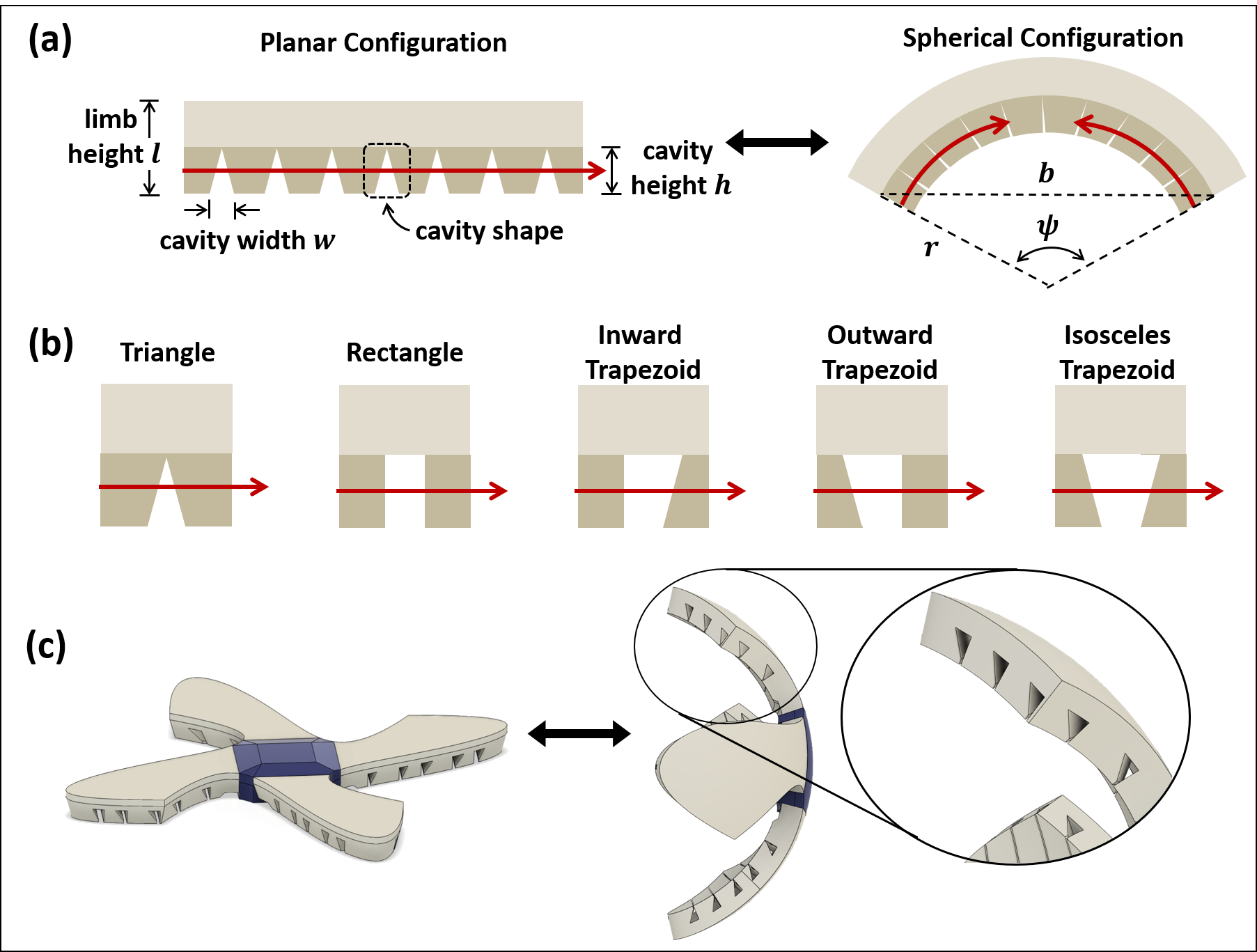}
\caption{The MTA limb is considered as a beam (red indicates tendon and arrow indicates the direction of actuation or pull). (a) Geometrically, an infinitesimal width rectangular cavity in the spherical configuration appears triangular in the planar configuration. The cavity width $w$ defines the maximum radius of the bend. (b) Five different cavity geometries are explored as the width $w$ is kept constant. (c) {The outward trapezoidal geometry cavity is used to design MSoRos that can physically reconfigure between the two configurations}}
\label{Fig:Cavitymodifications}
\end{center}
\end{figure}

Imagine uncurling of a MTA limb where an infinitesimal rectangular cavity in spherical configuration will result in a triangular shaped cavity in the planar configuration, Fig. \ref{Fig:Cavitymodifications}a. Considering the limb as an planar beam, the cavity width $w$ determines the bending curvature beyond which the MTA limb will experience high stiffness.  Geometrically, the desired cavity width $w_{desired}$ is
\begin{align}
w(b,r) &= \frac{r\psi - (r-h)\psi}{m} = \frac{2h}{m}\sin^{-1}\left(\frac{b}{2r}\right)\nonumber\\
w_{desired}&= \frac{2h}{m}\sin^{-1}\left(\frac{a}{2R}\right)
\end{align}
where $m,h$ are number of cavities and their height respectively. For the desired spherical configuration, $b,r$ are $a,R$ respectively (given the base platonic solid, Fig. \ref{Fig:PlatonicSolidsSpherical}). %
However, these calculations are purely geometric and the material properties are not considered. Interestingly, the manipulation of the cavity geometry can directly influence the limb stiffness without compromising the actuator's structural integrity. We consider five profiles (increasing order of polygon area) - triangle, rectangle, inward (right) trapezoid, outward (right) trapezoid and isosceles (inward-outward) trapezoid, Fig. \ref{Fig:Cavitymodifications}b. The increase in shape polygon area implies removal of soft material. The `outward' and `inward' labels correspond to how the right angle side of the trapezoids is aligned with the direction of actuation (direction of maroon arrow). The right angled side is `pulled' toward the non-right angle side for the inward trapezoid shape, and vice versa for the outward. 
\begin{figure*}
\begin{center}
\includegraphics[trim={5pt 5pt 30pt 5pt},clip,width=\textwidth]{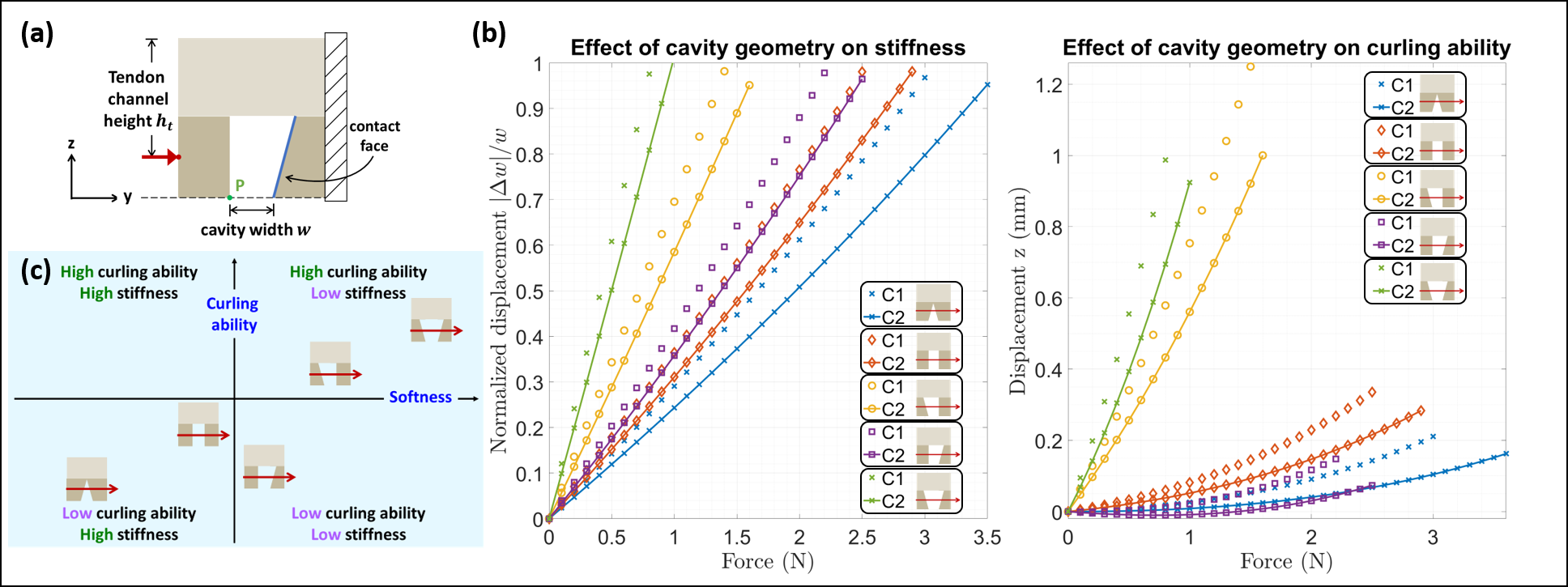}
\caption{{(a) The effect of cavity geometry analyzed by simulating a limb section with cavity width $w$ and a point force acting at height $h_t$ at the center of a surface while the other section face is assumed to be fixed. Two cases: (C1) $h_t=15mm$ - center of limb and (C2) $h_t=16.45mm$ - below limb center, are simulated.} (b) The movement of point $P$ is probed in $y,z$ direction. The `\textit{stiffness}' and the `\textit{curling ability}' are quantified as the relationships between normalized displacement and applied force, and the z-displacement and applied force, respectively. Steeper slopes imply more softness and higher curling ability. (c) The curling ability to softness plot illustrates the various regions including that of high curling ability and low stiffness - outward and isosceles trapezoid cavity geometry.}
\label{Fig:Cavitygeometry}
\end{center}
\end{figure*}
 
The limb stiffness and curling ability is analyzed by simulating the deformation of the limb section with single cavity for different geometries.  The simulations are performed in Autodesk Fusion 360\textregistered~ using non-linear static studies with customizable material properties for elastomers. Empirically determined Mooney-Rivlin constants are used for hyperelastic Smooth-on Dragon Skin\texttrademark 10A \cite{sasso_characterization_2008,xavier_finite_nodate}. The response to motor-tendon actuation is estimated by simulating a single cross-section of the limb where the tendon force is assumed to be a point force acting normal to the surface at a distance $h_t$ from the top surface at the middle of the face while the other side of the limb section is constrained not to move, Fig. \ref{Fig:Cavitygeometry}a. We probe the movement of the point $P$ at the tip of the cavity for displacements in $y$ and $z$ directions. The simulation model considers limb height to be $l=30mm,h=20mm,w=2.35mm$ and trapezoid obtuse angle to be $70\deg$, and two cases: (C1) $h_t=15mm$, (C2) $h_t=16.45mm$. 

The stiffness of the limb section is quantified as the relationship between the normalized displacement in $y$ direction, $\Delta w/w$, and the applied force. Similarly, the relationship between the $z$ displacement and the force is defined as the curling ability of the section, Fig. \ref{Fig:Cavitygeometry}b. The results re-affirm the intuitive hypothesis that increase in the cavity polygon area will decrease stiffness and the material will feel softer. {However}, the effect of geometry on the curling ability is not intuitive due to the directional nature of the force. The behavior of inward and outward trapezoid geometries is starkly visible in their curling ability, where the former is out-performed by the latter.  The isosceles trapezoid geometry is seen to be the most soft with highest curling ability. The simulation results present the outward and isosceles trapezoid geometries to be most suitable for MSoRos that reconfigure between a planar and spherical configuration, Fig. \ref{Fig:Cavitygeometry}c. Experimentally, for the chosen MTA, it is observed that outward trapezoid geometry provides the optimal stiffness and curling ability to lift the weight of the electronic payload and transform between the two configurations. Additionally, the experiments highlight the uncurling ability of inward and isosceles trapezoid geometries is hindered due to the frictional interactions when the point $P$ moves along the contact face. The MSoRs are fabricated using the outward trapezoid cavity geometry, Fig. \ref{Fig:Cavitymodifications}c. The curling-uncurling is achieved through actuation of the MTA. However, the Gaussian curvature of the curled limb surface (product of the principle curvatures) depends upon the routing path of the tendon inside the limb. This analysis with regard to the placement of the MTA is outside the scope of the paper.

\section{Fabrication and Spherical Reconfiguration}\label{Sec:Fabrication}
\begin{figure}
\begin{center}
\includegraphics[width=0.65\columnwidth]{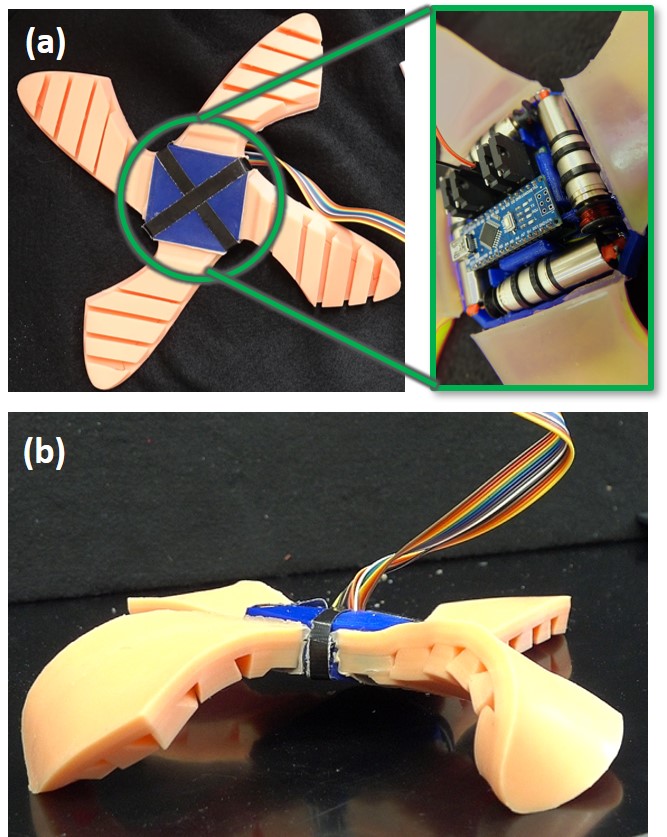}
\caption{(a) Fabricated modular soft robot with four limbs. The motor and spool are enclosed inside the flexible 3D printed blue hub. The views illustrate the modified fin design and (b) the MSoRo when two of the four MTAs are actuated.}
\label{Fig:FabricateModules}
\end{center}
\end{figure}

A four limb spherically reconfigurable MSoRo capable of planar locomotion is designed by using a cube of {$a=11cm$} edge (four edges and six faces) as the base platonic solid and module topology curve of sinusoidal family with {$A=0.86$}. The MSoRo comprises four MTA limbs and a central hub. The central hub is 3D printed using NinjaFlex\textregistered~ thermoplastic polyurethane, and houses the control and actuation payload (motor and electronics), Fig. \ref{Fig:FabricateModules}. Each limb of height $l=30mm$ is cast in planar configuration using Smooth-On Dragon Skin\texttrademark~ silicone rubber (mixing liquid silicone components and degassing in vacuo) with {$m=5$} outward trapezoidal geometry cavities ($h=20mm$, trapezoid obtuse angle $70\deg$) that allow them to curl between the planar and spherical configuration with curvature radius of $R=19cm$. Rapid curling and uncurling of the soft limbs is achieved using MTA that comprises of a backdrivable motor, pulley, fishing line tendon ($h_t=16.45mm$) and treble hook for anchoring. The azimuthal projection between the two configurations implies increase of distortion with the radial distance from the center of projection. This informs the fabrication and limb design decisions in a three-fold manner. Firstly, the central hub experiences minimal distortion during  reconfiguration, permitting encapsulation of the stiffest robot components at this position. Secondly, the stiffness theoretically decreases with the radial distance, facilitating realization of equal Gaussian curvatures along the principal axes (sphere) when multiple modules dock in the spherical configuration. This will be possible due to the intermodular interaction forces that arise during docking. Finally, the MSoRo can be cast in a planar configuration which allows for passive uncurling of the limb that is assisted by the elastic energy stored during curling.

\begin{figure}
\begin{center}
\includegraphics[width=0.9\columnwidth]{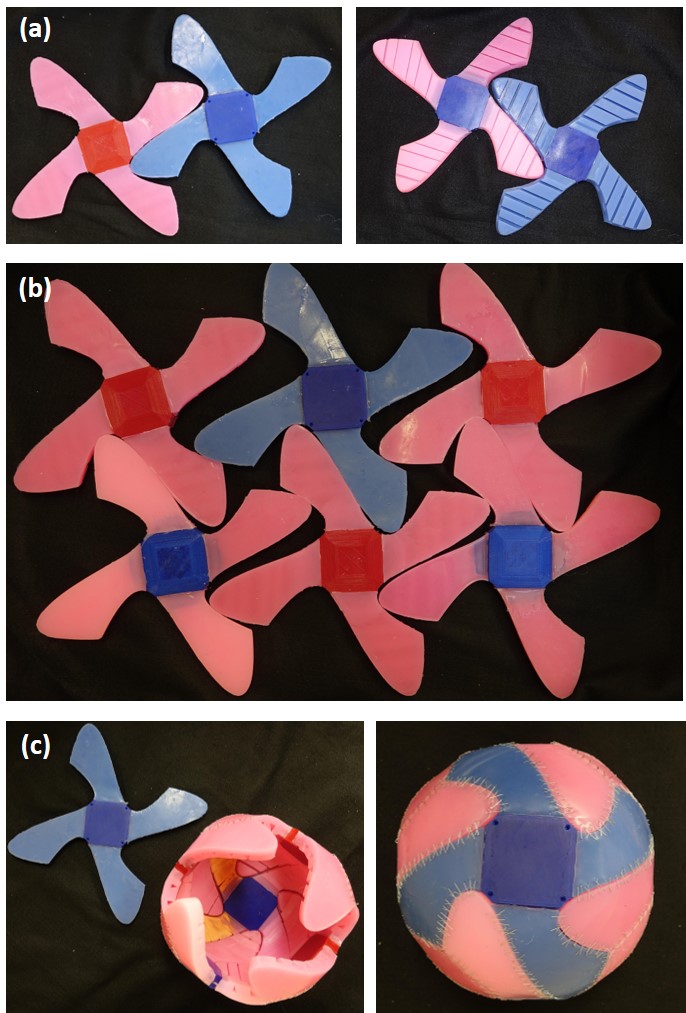}
\caption{(a) Four-limb MSoRos with cube as the base platonic solid are fabricated using silicone rubber and MTA actuators. The `outward trapezoid' geometry of the limb cavities allow them to reconfigure between planar and spherical configurations. (b) Six identical modules in planar configuration. The `gaps' between them highlight the distortions that occur due to sphere-to-plane distortion (intramodular distortions). (c) The six robot modules in spherical configuration assemble into a sphere.}
\label{Fig:FabricatedShapeMorphing}
\end{center}
\end{figure}

The Fig. \ref{Fig:FabricatedShapeMorphing} illustrates the four-limb MSoRos on a planar surface and the gaps (distortion) between them when projected on flat surface. Collectively, the six fabricated MSoRos can be reconfigured into a sphere which has different dynamics in comparison to the planar configuration. These modules do not have docking mechanisms and are mechanically connected for the experimental validation. The multimedia attachment illustrates four-limb MSoRos reconfigured into a sphere and performing locomotion (translation and  rotation) in the planar configuration. The supplemental video illustrates four-limb MSoRos reconfigured into a sphere and performing locomotion (translation and  rotation) in the planar configuration. The fabrication and actuation of the MSoRos serve as experimental validation for the topology and morphology design concepts presented in this paper. The fabricated MSoRos integrate multiple components for different stiffness - rigid motors and electrical components, flexible hub and the soft body. Empirical adjustments to the design (e.g., cavity spacing) have been made to compensate for the material non-uniformity. The determination and analysis of these adjustments for the multi-material and multi-component robot are considered outside the scope of this paper.
